\documentclass[conference]{IEEEtran}

\IEEEoverridecommandlockouts
\usepackage{cite}
\usepackage{amsmath,amssymb,amsfonts}
\usepackage{algorithmic}
\usepackage{graphicx}
\usepackage{textcomp}
\usepackage{xcolor}
\def\BibTeX{{\rm B\kern-.05em{\sc i\kern-.025em b}\kern-.08em
    T\kern-.1667em\lower.7ex\hbox{E}\kern-.125emX}}
    
\usepackage[T2A,T1]{fontenc}
\usepackage[utf8]{inputenc} 
\usepackage[russian,english]{babel} 

\usepackage{mdframed}
\usepackage{listings}
\usepackage{hyperref}

\begin{document}

\title{Adapting the LodView RDF Browser for \\ Navigation over the Multilingual \\ Linguistic Linked Open Data Cloud

\thanks{
\hskip -1em \rule{\columnwidth}{0.4pt} \\
\hskip -1em Please, cite as: Alexander Kirillovich and Konstantin Nikolaev. Adapting the LodView RDF Browser for Navigation over the Multilingual Linguistic Linked Open Data Cloud // Proceedings of the 9th IEEE International Conference on Sciences of Electronics, Technologies of Information and Telecommunications (SETIT 2022), Genova, Italy \& Sfax, Tunisia, 28-30 May 2022. IEEE, 2022. Pp. 143--149 [revised version]. \url{https://arxiv.org/abs/2208.13295}
}
}

\author{\IEEEauthorblockN{Alexander Kirillovich}
\IEEEauthorblockA{
\textit{Kazan Federal University},\\
\textit{Joint Supercomputer Center}\\
\textit{of the Russian Academy of Sciences}\\
Kazan, Russia \\
alik.kirillovich@gmail.com}
\and
\IEEEauthorblockN{Konstantin Nikolaev}
\IEEEauthorblockA{
\textit{Kazan Federal University},\\
\textit{Joint Supercomputer Center}\\
\textit{of the Russian Academy of Sciences}\\
Kazan, Russia \\
konnikolaeff@yandex.ru}
}

\maketitle

\thispagestyle{plain}
\pagestyle{plain}
\setcounter{page}{143}

\begin{abstract}
The paper is dedicated to the use of LodView for navigation over the multilingual Linguistic Linked Open Data cloud. First, we define the class of Pubby-like tools, that LodView belongs to, and clarify the relation of this class to the classes of URI dereferenciation tools, RDF browsers and LOD visualization tools. Second, we reveal several limitations of LodView that impede its use for the designated purpose, and propose improvements to be made for fixing these limitations. These improvements are: 1) resolution of Cyrillic URIs; 2) decoding Cyrillic URIs in Turtle representations of resources; 3) support of Cyrillic literals; 4) user-friendly URLs for RDF representations of resources; 5) support of hash URIs; 6) expanding nested resources; 7) support of RDF collections; 8) pagination of resource property values; and 9) support of \LaTeX\ math notation. Third, we partially implement several of the proposed improvements.
\end{abstract}

\begin{IEEEkeywords}
LodView, Pubby-class tool, RDF browser, dereferenciation tool, Linked Data visualization, Linguistic LOD
\end{IEEEkeywords}
%%%% BEGIN COPYRIGHT NOTICE
\makeatletter
\def\ps@IEEEtitlepagestyle{
  \def\@oddfoot{\mycopyrightnotice}
  \def\@evenfoot{}
}
\def\mycopyrightnotice{
  {\footnotesize 978-1-7281-8442-5/22/\$31 ~\copyright~2022 IEEE\hfill} 
  \gdef\mycopyrightnotice{}
}
\makeatother
%%%% END COPYRIGHT NOTICE
\section{Introduction}
Digital language resources are key to linguistics research and natural language processing. Among these resources are corpora, thesauri, lexicons, grammatical dictionaries, valence dictionaries, linguistic data categories registers, and other. 

Publishing linguistic resources according to the principles of Linked Open Data (LOD) \cite{bizer2009, heath2011, hogan2020} is a recently established trend, that has led to the emergence of the fast-growing Linguistic Linked Open Data (LLOD) cloud \cite{mccrae2016,cimiano2020}. It offers a number of advantages, including structural and conceptual interoperability, and integration of heterogeneous linguistic resources to solve a common problem.

As a part of the global LLOD project we are working on developing the fragment of the LLOD cloud for Russian and minority languages of Russia. This fragment integrates several resources such as the RuThes \cite{kirillovich2017}, TatThes \cite{galieva2017}, TatWordNet~\cite{kirillovich2021}, TatVerbBank, the Russian corpus ``OpenCorpora'' \cite{kirillovich2022}, the Tatar national corpus ``Tugan Tel'' \cite{mukhamedshin2020OSS, mukhamedshin2020SETIT, nevzorova2016}, the parallel informal/formal corpus of educational mathematical texts in Russian \cite{kirillovich2020CSDEIS}, the linguistic layer of the educational mathematical ontology OntoMath\textsuperscript{Edu} \cite{Kirillovich2019,Kirillovich2020,Kirillovich2021WEA} and other. 

In accordance with Linked Open Data best practices, the integrated resources have to be available via RDF dumps, SPARQL endpoints and dereferenceable URIs. For navigation over the LOD cloud via dereferenceable URIs, a specialized tool is required. One of the most popular navigation tools is the LodView RDF browser.

\begin{figure}[t]
\includegraphics[width=\columnwidth]{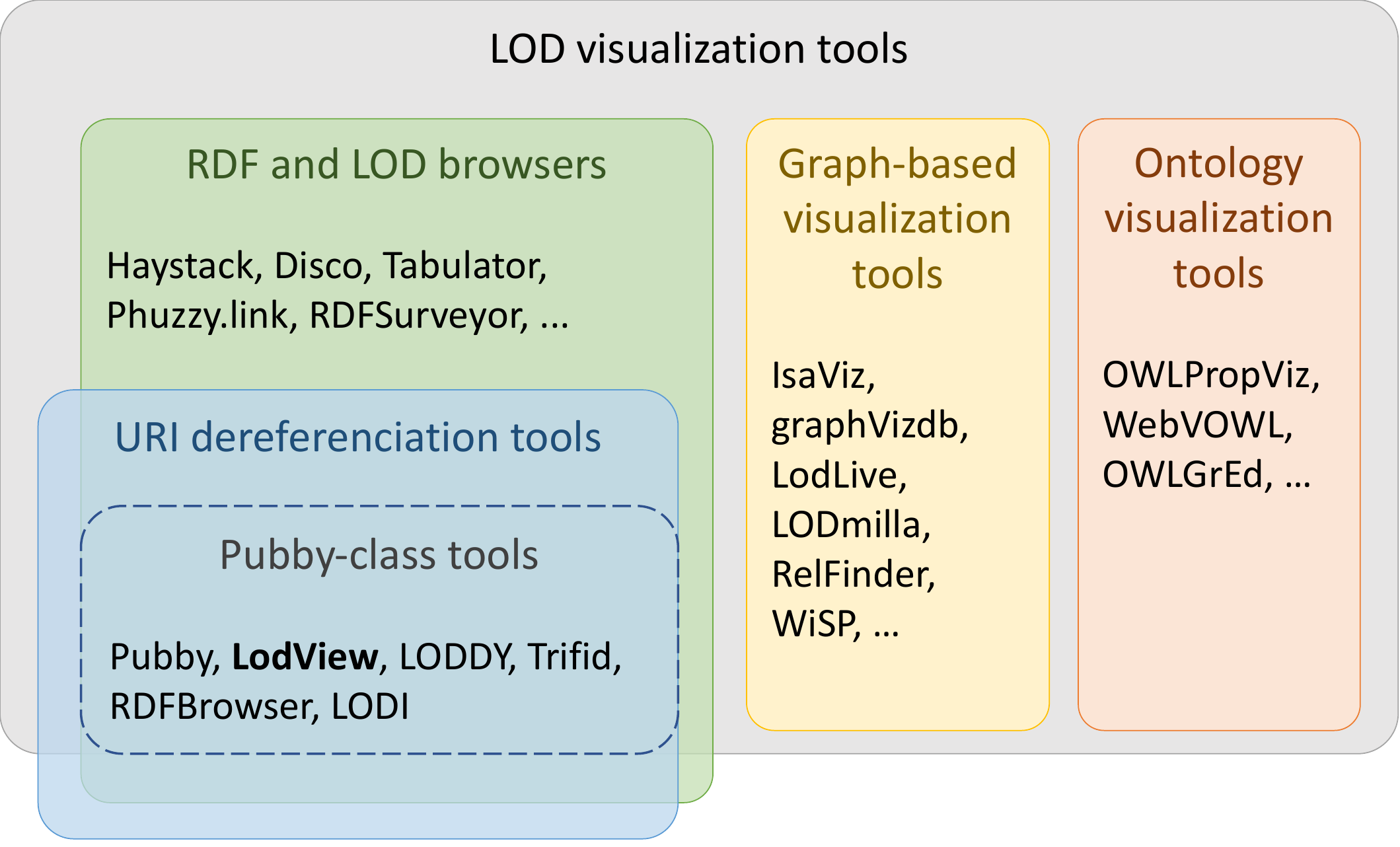}
\caption{The place of Pubby-class tools among other LOD applications.} \label{fig:related-works}
\end{figure}

\begin{figure*}
    \centering
    \includegraphics[width=0.8\textwidth]{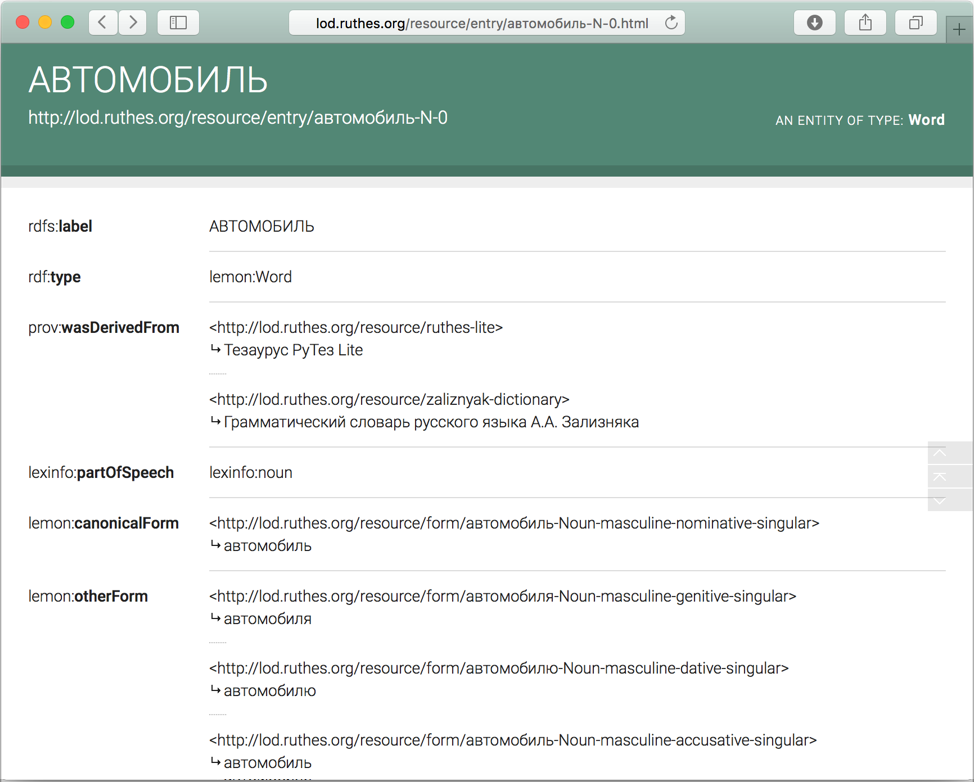}
    \caption{Visualization of the ‘\textcyrillic{автомобиль}’ (`automobile') lexical entry in LodView} \label{fig:lodview}
\end{figure*}

LodView (\url{https://github.com/dvcama/LodView}) is a web application for dereferencing URIs of the RDF resources stored in the SPARQL-exposed datasets. When a human user requests the URI of an RDF resource, LodView retrieves information of the resource in the dataset and visualizes this information as a human-readable HTML page. On the page, the resource properties are represented in a tabular form. Each row represents a property-value pair; the first cell of the row contains the property name and the second contains the property value or values. If a property value is the URI of an another resource, this value is rendered as a link. A user may follow the link and get representation of the another resource, thus navigating over RDF dataset. The HTML page also contains links to the RDF documents with machine-readable representation of the resource (in different serialization format, including N-Triples, Turtle, RDF/XML and JSON-LD). When the URI is requested not by human but by a software agent, LodView immediately returns the machine-readable RDF representation.

Fig. \ref{fig:related-works}. shows the place of LodView among other LOD applications.

Fig. \ref{fig:lodview} represents an example how LodView visualizes the ‘\textcyrillic{автомобиль}’ (`automobile')  lexical entry.

However, LodView has several limitations that impede its use for navigation over the multilingual LLOD cloud. In this paper, we continue the work initiated in \cite{kirillovich2020itnou,Kirillovich2020DAMDID} and present the first version of our LodView fork, adapted to address these limitations.

The rest of the paper is organized as following. In Section~2 we define the class of Pubby-like tools that LodView belongs to and outline the relationships of this class with other classes of LOD applications. In Section 3 we list the limitations of LodViews and propose improvements to be made for fixing these limitations. Section~4 describes the improvements, that have already been made by us. In Conclusion, we outline the directions of future work.

\section{Related Works}

In this section we clarify the place of LodView among other LOD applications.

%In this section we define the class of tools that LodView belongs to and outline the relationships of this class with other classes.

LodView belongs to a class of tools, that we called the Pubby class after its first prominent representative. We define a Pubby-class tool as a LOD application which is:

\begin{enumerate}

\item a URI dereferenciation tool, i.e. when a client requests the URI of an RDF resource via HTTP protocol, the tool handles the request and responds with a resource representation;

\item a LOD visualization tool, i.e. it represents the resource in human-friendly visual form;

\item  a Linked Data or RDF browser, i.e. it allows a user to navigate between RDF resources of an RDF dataset or the entire LOD cloud, by following links in resource representations;

\item and a SPARQL endpoint interface, i.e. it retrieves  information about resources from a SPARQL-exposed dataset.
\end{enumerate}

Conceptually, these four classes are independent and so it is possible for example to conceive a URI dereferenciation tool that doesn't allow a user to follow links or even doesn't visualize a resource at all, dereferencing it just as a source RDF file in some textual serialization format. However, in practice, every known URI dereferenciation tool is an RDF browser, and every known RDF browser is a LOD visualization tool (see the relationships between the classes at Fig. \ref{fig:related-works}).

\textbf{Pubby-class tools.} Besides LodView, the Pubby class includes the following tools:

\begin{itemize}
\item Pubby\footnote{\url{http://wifo5-03.informatik.uni-mannheim.de/pubby/}}. A Java web application. Was used for publishing a lot of LOD datasets, including DBpedia. However, it hadn't been updated in more then 10 years and now its GitHub repository is officially archived.

\item LODDY\footnote{\url{https://bitbucket.org/art-uniroma2/loddy/src/master/}}. A Java web application. Is used for publishing the AGROVOC Linked Dataset.

\item Zazuko Trifid\footnote{\url{https://github.com/zazuko/trifid}}. A Node.js-based web application. Is used in two open data platforms of Swiss Confederation.

\item RDFBrowser\footnote{\url{https://github.com/okgreece/RDFBrowser}}. A PHP web application. Is used by the OpenBudgets.eu platform.

\item LODI\footnote{\url{https://github.com/marfersel/LODI}}. A Node.js-based web application. Is used by two Spanish open data portals. %, Open Data Cáceres and Open Data UNEx.
\end{itemize}

% While almost all Pubby-class tools are Linked Data browsers, there are Linked Data browsers that are not Pubby-class tools. The difference between Pubby-class and not-Pubby-class Linked Data browsers is in the URI dereferenciation mechanism. An instance of Pubby-class tool runs on a web server working on a particular domain and serves those RDF resources, whose URIs belong to the namespace that this server handles (or are mapped to by a predefined mapping). To get representation of a such resource, a user just types its URI in the browser's address bar. In contrast, an instance of not-Pubby-class Linked Data browser provides access to information about resources regardless of the namespace their URIs belongs to. To get representation of a resource by this tool, a user can't just type its URI in the browser's address bar. Instead, he or she should at first open the tool, and then type the URI in a custom field in the tool UI.

\textbf{Other Linked Data and RDF browsers.} While almost all Pubby-class tools are Linked Data browsers, there are Linked Data browsers that are not Pubby-class tools. Among the first such Linked Data browsers the well-known ones are Haystack \cite{quan2004}, Disco\footnote{\url{http://wifo5-03.informatik.uni-mannheim.de/bizer/ng4j/disco/}} and Tabulator\footnote{\url{https://github.com/linkeddata/tabulator}} \cite{bernerslee2006}. Examples of the more recent developments are Phuzzy.link\footnote{\url{https://github.com/blake-regalia/phuzzy.link}} \cite{blake2017} and RDFSurveyor\footnote{\url{https://github.com/guiveg/rdfsurveyor}} \cite{vega-gorgojo2019}.

Besides Linked Data browsers, there are many LOD visualization tools of several other types, including graph-based and ontology visualization tools.

\textbf{Graph-based visualization tools.} While RDF browsers mainly work on the level of single RDF resources, graph-based tools work on the level of entire RDF graphs. One of the earliest tools in this category is IsaViz\footnote{\url{https://www.w3.org/2001/11/IsaViz/}}, a desktop application for authoring and visualizing RDF models, represented as relatively small local RDF files. In contrast, graphVizdb\footnote{\url{http://www.graphvizdb.com/}} \cite{bikakis2016} enables the user to navigate over very-large RDF graphs. LodLive\footnote{\url{https://github.com/LodLive/LodLive}} \cite{camarda2012} and LODmilla\footnote{\url{https://github.com/dsd-sztaki-hu/LODmilla-frontend}} \cite{micsik2014,micsik2015} provide incremental graph visualization. RelFinder\footnote{\url{http://www.visualdataweb.org/relfinder.php}} \cite{heim2010} and WiSP\footnote{\url{https://github.com/GTartari/Weighted-Shortest-Paths}} \cite{tartari2018} are intended to identify and visualize the paths between two selected RDF resources.

\textbf{Ontology visualization tools.} Examples of these tools are OWLPropViz, WebVOWL\footnote{\url{https://github.com/VisualDataWeb/WebVOWL}} \cite{lohmann2014} and OWLGrEd\footnote{\url{http://owlgred.lumii.lv/}} \cite{liepins2014}.

For a book length overview of LOD visualization tools, we refer the interested reader to \cite{po2020}.

\section{Required improvements of LodView}

In this section, we list the limitations of LodViews and propose improvements to be made to fix these limitations.

\begin{figure}[h]
\frame{\includegraphics[width=\columnwidth]{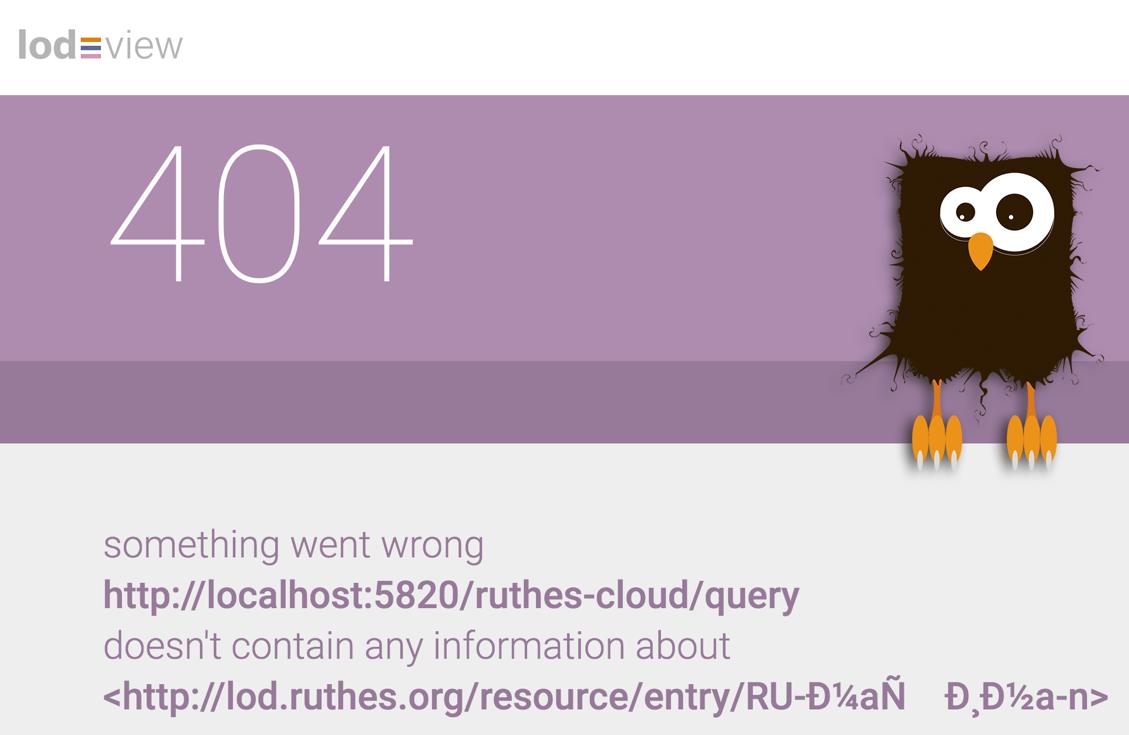}}
\caption{Error occurred while resolving a URI that contains Cyrillic characters.} \label{fig:resolution-of-cyrillic-uris}
\end{figure}

\textbf{1) Resolution of Cyrillic URIs.} In a LLOD dataset, the URI of an RDF resource may contain Cyrillic characters. For example, in the RuThes Cloud dataset, the URI of the lexical entry ‘\textcyrillic{машина}’ (`car') is <http://lod.ruthes.org/resource/entry/RU-\textcyrillic{мaшинa}-n>. However, in some environments, resolving Cyrillic URIs by LodView leads to an error (see Fig. \ref{fig:resolution-of-cyrillic-uris}). LodView must resolve Cyrillic URIs in all environments.

\begin{figure}[ht]
\includegraphics[width=\columnwidth]{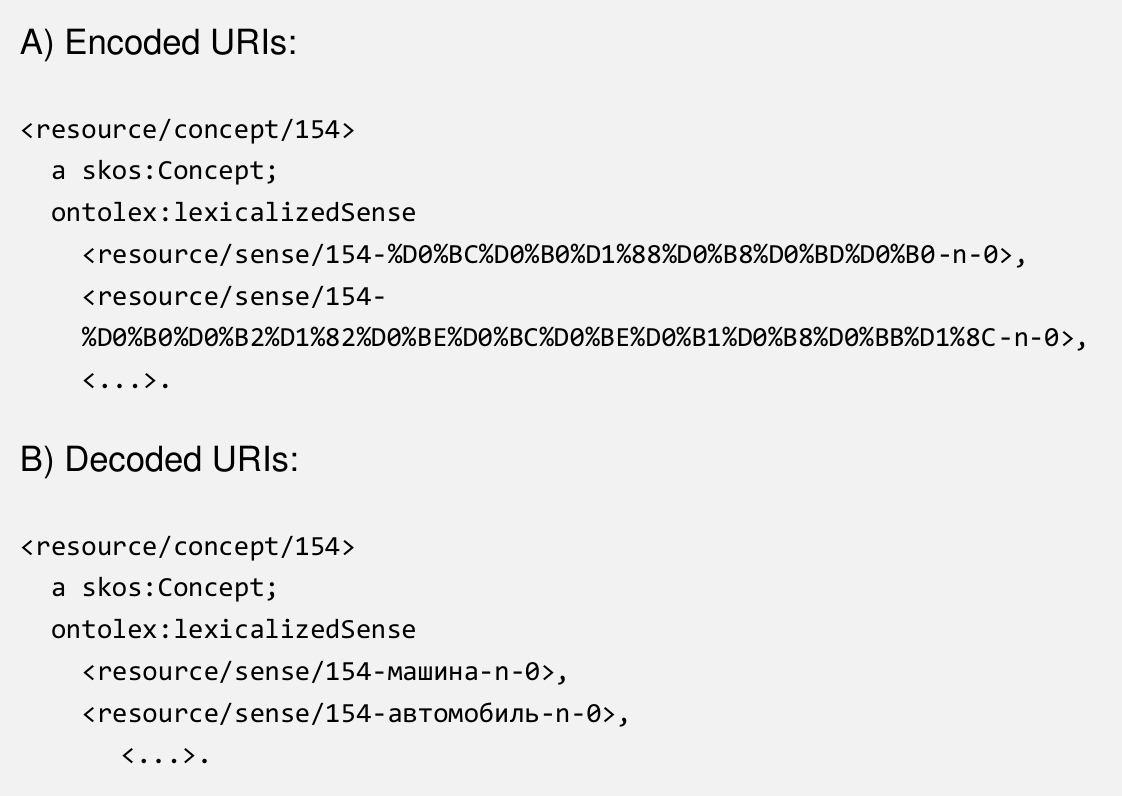}
\caption{Fragment of a Turtle representation of a resource with encoded and decoded URIs.} \label{fig:cyrillic-uris}
\end{figure}

\textbf{2) Decoding Cyrillic URIs in Turtle representations of resources.} In a machine-readable RDF representation of a resource in the Turtle serialization format, Cyrillic URIs are presented in an encoded form. Fig. \ref{fig:cyrillic-uris}A shows a fragment of the Turtle representation of the \textit{Automobile} concept, that contains encoded URIs. Encoded URIs hinder readability of Turtle documents. In the Turtle representation of a resource, URIs must be presented in a decoded form (for example, as in Fig. \ref{fig:cyrillic-uris}B).

\begin{figure}[ht]
\includegraphics[width=\columnwidth]{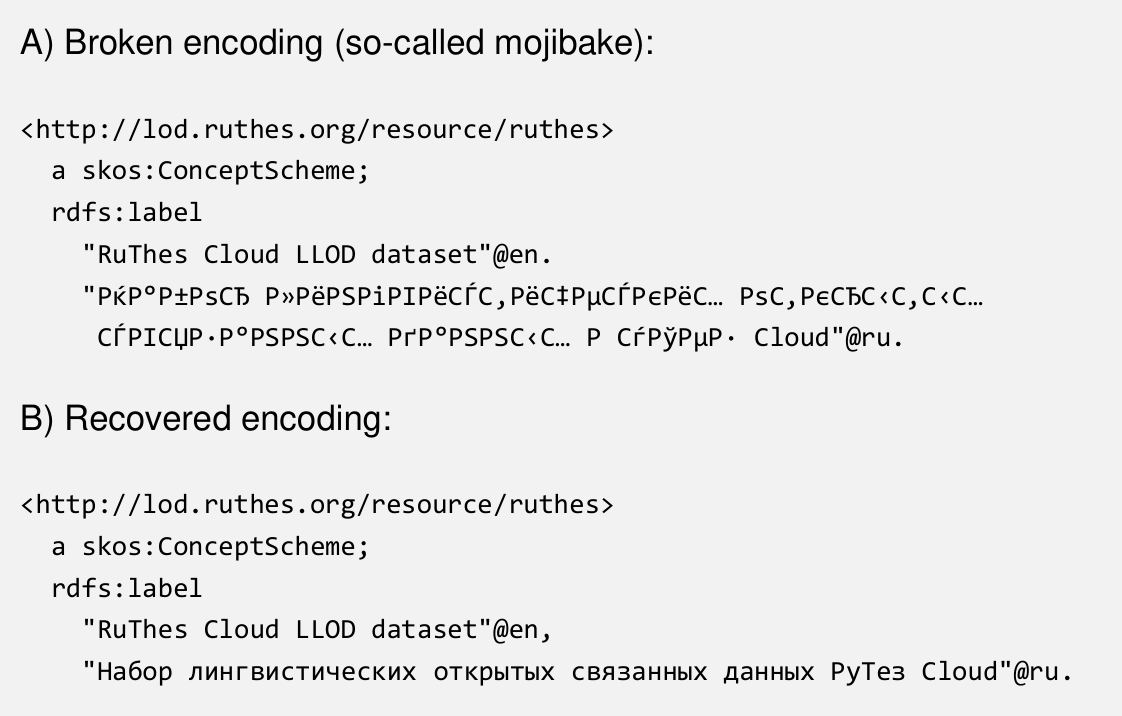}
\caption{Fragment of a Turtle representation of a resource with broken and recovered literals.} \label{fig:cyrillic-literals}
\end{figure}

\textbf{3) Support of Cyrillic literals.} In a machine-readable RDF representation of a resource, Cyrillic literals are in broken encoding. This happens even if the dataset was encoded by Unicode. Fig. \ref{fig:cyrillic-literals}A shows a Turtle representation of the RuThes Cloud resource, that contains literals in broken encoding. LodView must represent Cyrillic literals correctly (as in Fig. \ref{fig:cyrillic-literals}B).

\textbf{4) User-friendly URLs for RDF representations of resources.} LodView resolves URI’s of RDF resources in accordance with the content negotiation principle. When the URI of an RDF resource is requested on by a human user, LodView redirects the user to the URL of the HTML page with human-readable resource representation. When the URI is requested by a software agent, LodView redirects the agent to the URL of RDF document with the machine-readable representation.

% URLs of HTML pages are of user-friendly format: \texttt{<RESOURCE\_URI.html>}. For example, the resource identified by the URI \texttt{<http://lod.ruthes.org/ resource/entry/RU-\textcyrillic{мaшинa}-n>} is represented by the HTML document located at the URL \texttt{<http://lod.ruthes.org/resource/entry/RU- \textcyrillic{мaшинa}-n.html>}. However, URLs of RDF documents are rather cumbersome. For example, the Turtle representation of the aforementioned resource is located at the URI \texttt{<http://lod.ruthes.org/resource/entry/ RU-\textcyrillic{мaшинa}-n?output=appli-cation\%2Frdf\%2Bxml>}. In order the URL’s of RDF documents can be easy to read, they must be of user-friendly formats: \texttt{<RESOURCE\_URI.ttl>}, \texttt{<RESOURCE\_URI.n3>}, etc. For example, the Turtle representation of the \texttt{<http://lod.ruthes.org/ resource/entry/RU-\textcyrillic{мaшинa}-n>} resource must be assigned the URI \texttt{<http://lod.ruthes.org/ resource/entry/RU-\textcyrillic{мaшинa}-n.ttl>}.

URLs of HTML pages are of user-friendly format: <RESOURCE\_URI.html>. For example, the resource identified by the URI <http://lod.ruthes.org/resource/entry/RU-\textcyrillic{мaшинa}-n> is represented by the HTML document located at the URL <http://lod.ruthes.org/resource/entry/RU-\textcyrillic{мaшинa}-n.html>. However, URLs of RDF documents are rather cumbersome. For example, the Turtle representation of the aforementioned resource is located at the URI <http://lod.ruthes.org/resource/entry/RU-\textcyrillic{мaшинa}-n?output=appli-cation\%2Frdf\%2Bxml>. In order the URL’s of RDF documents can be easy to read, they must be of user-friendly formats: <RESOURCE\_URI.ttl>, <RESOURCE\_URI.n3>, etc. For example, the Turtle representation of the <http://lod.ruthes.org/resource/entry/RU-\textcyrillic{мaшинa}-n> resource must be assigned the URI <http://lod.ruthes.org/resource/entry/RU-\textcyrillic{мaшинa}-n.ttl>.

\textbf{5) Support of hash URIs}. In LLOD datasets, RDF resources may be identified by hash URIs. A hash URI is a URI, that contains a special part separated from the base part of the URI by a hash symbol (\#). This special part is called the ‘fragment identifier’. Usually, a hash URI identify a resource, that is somehow subordinated to the resource identified by the base part of this URI. For example, in the Princeton WordNet dataset, the hash URI <http://wordnet-rdf.princeton.edu/wn31/cat-n\#CanonicalForm> identifies the canonical form of the lexical entry ‘cat’, while the lexical entry itself is identified by the base part <http://wordnet-rdf.princeton.edu/wn31/cat-n>. In this case, the URI of the canonical form of the lexical unit is formed by the URI of this lexical entry concatenated with the \texttt{\#CanonicalForm} fragment identifier.

However, LodView can’t resolve URIs with fragment identifiers. So, for example, when a user try to retrieve the hash URI <http://wordnet-rdf.princeton.edu/wn31/cat-n\#CanonicalForm>, LodView sends a representation not of the resource identified by this URI, but of the resource identified by the URI <http://wordnet-rdf.princeton.edu/wn31/cat-n>. This problem is caused by the HTTP protocol: when a client wants to retrieve a hash URI, the HTTP protocol requires the fragment part to be stripped off before requesting the URI from the server.  This problem can be solved as follows: when LodView resolves a URI, its response must contain not only information of the requested resource, but also of all other resources identified by hash URIs with the same base part. For example, the representation of the resource identified by <http://wordnet-rdf.princeton.edu/wn31/cat-n> must contain information of this resource along with information of the resource identified by the hash URI <http://wordnet-rdf.princeton.edu/wn31/cat-n\#CanonicalForm>.

\begin{figure}[h]
\frame{\includegraphics[width=\columnwidth]{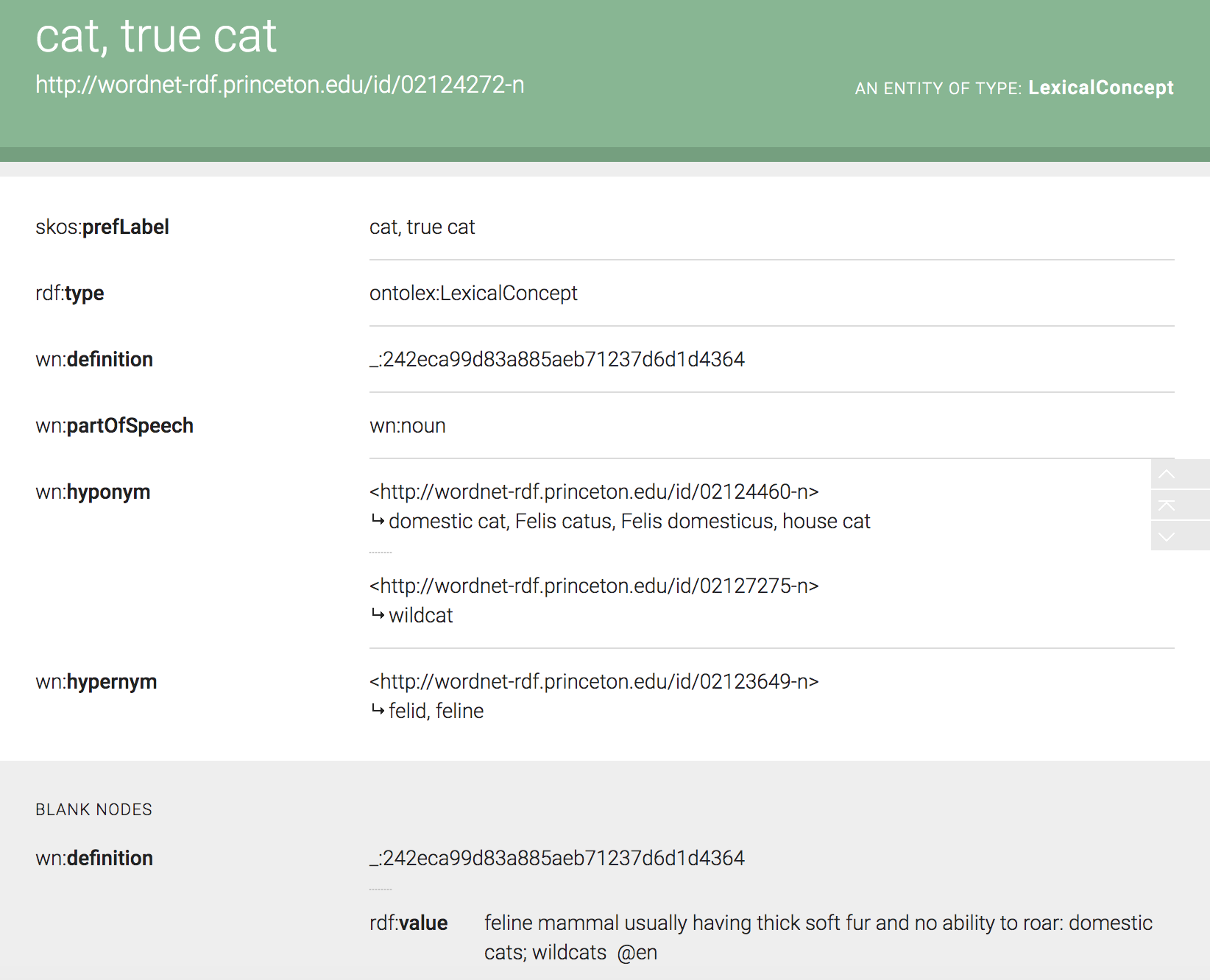}}
\caption{Blank node, located at the bottom of the page.} \label{fig:blank-node}
\end{figure}

\textbf{6) Expanding nested resources.} In LodView, the HTML representation of an RDF resource contains the table with the resource properties. In each row, the first cell contains the property name, and the second cell contains its value. The value of a property may be the URI of another, nested, resource. For the user, the only way to retrieve information about this nested resource is to follow the link and go to another HTML page. This need to leave the current page may spoil the user experience. LodView must allow the user to expand nested resources enabling him or her to see the properties of these resources without leaving the current page.

When the property value is a blank node, this value is represented in the table by the surrogate blank node ID. The description of this blank node itself is located at the bottom of the page. See for example Fig. \ref{fig:blank-node}, where a blank node is a value of the \texttt{wn:definition} property. In order to see information of the blank node, a user have to scroll the page. Again, LodView must allow the user to expand nested blank nodes to see its properties in place without scrolling the page. 

\begin{figure}[th]
\frame{\includegraphics[width=\columnwidth]{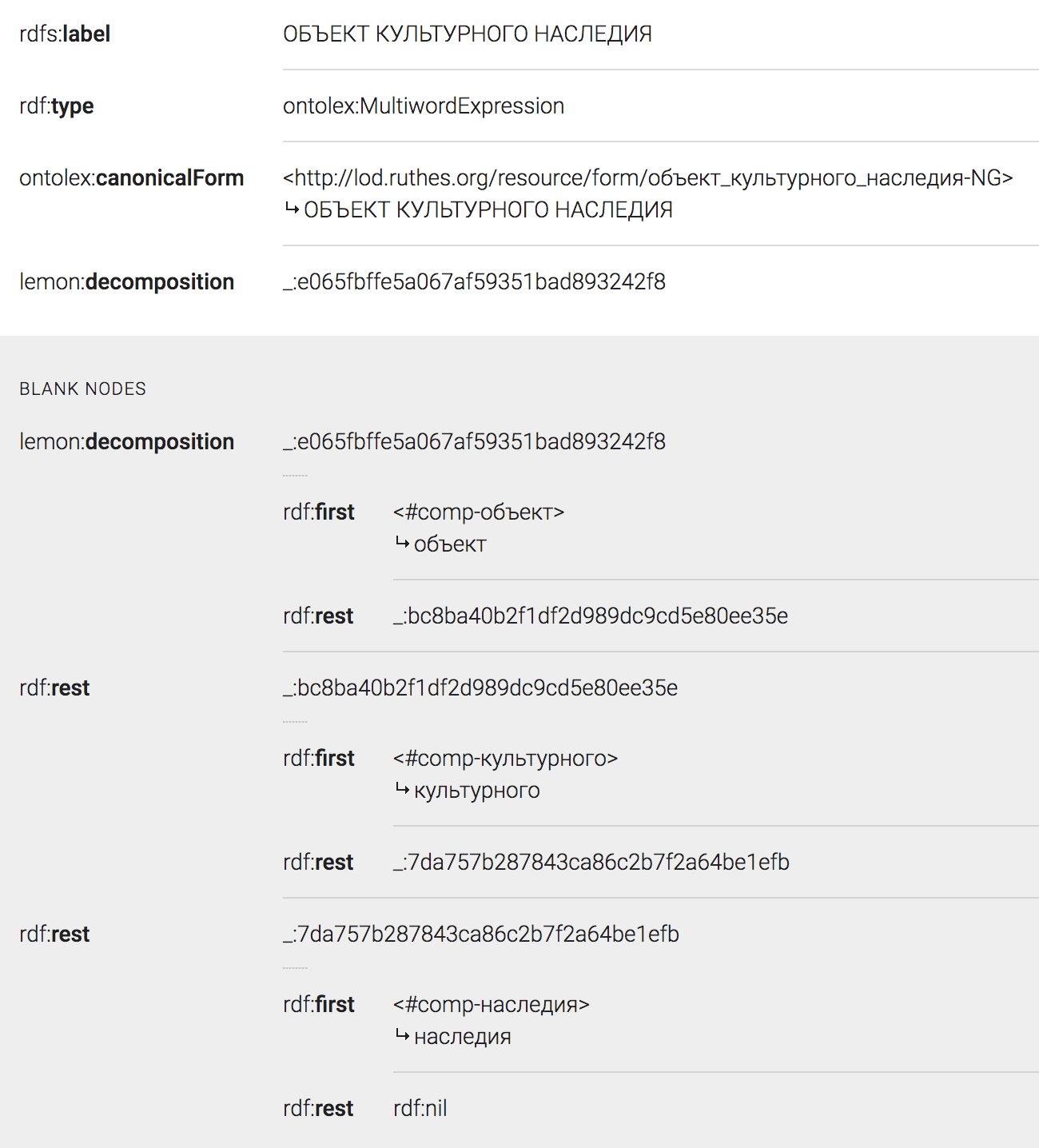}}
\caption{Visualization of an RDF collection.} \label{fig:rdf-list}
\end{figure}

\textbf{7) Support of RDF collections.} LLOD datasets may contain RDF collections. According to the RDF standard, a collection is represented as a low-level list structure in the RDF graph. When LodView visualize an RDF collection, it visualizes not the collection itself, but the underlying list structure. This way of visualizing makes comprehension of the collection difficult. An example of visualization of the RDF list \texttt{< <\#comp-\textcyrillic{объект}>, <\#comp-\textcyrillic{культурного}>, <\#comp-\textcyrillic{наследия}> >} is represented at Fig \ref{fig:rdf-list}. LodView must visualize RDF collections as a high-level sequence of its members.

\textbf{8) Pagination of resource property values.} An RDF resource may have an arbitrary number of values for a particular property. And this number can be very big. For example, the RuWordNet thesaurus is linked by the \texttt{lime:entry} property with more than 100 thousand lexical entries. However, LodView tries to display all these values at once, which can overwhelm the user with the amount of data, significantly slow down page loading and even cause server overload. To prevent these negative consequences, LodView must limit the number of the displayed property values by providing a pagination widget or just a ``Load more'' button.

\begin{figure}[t]
\frame{\includegraphics[width=\columnwidth]{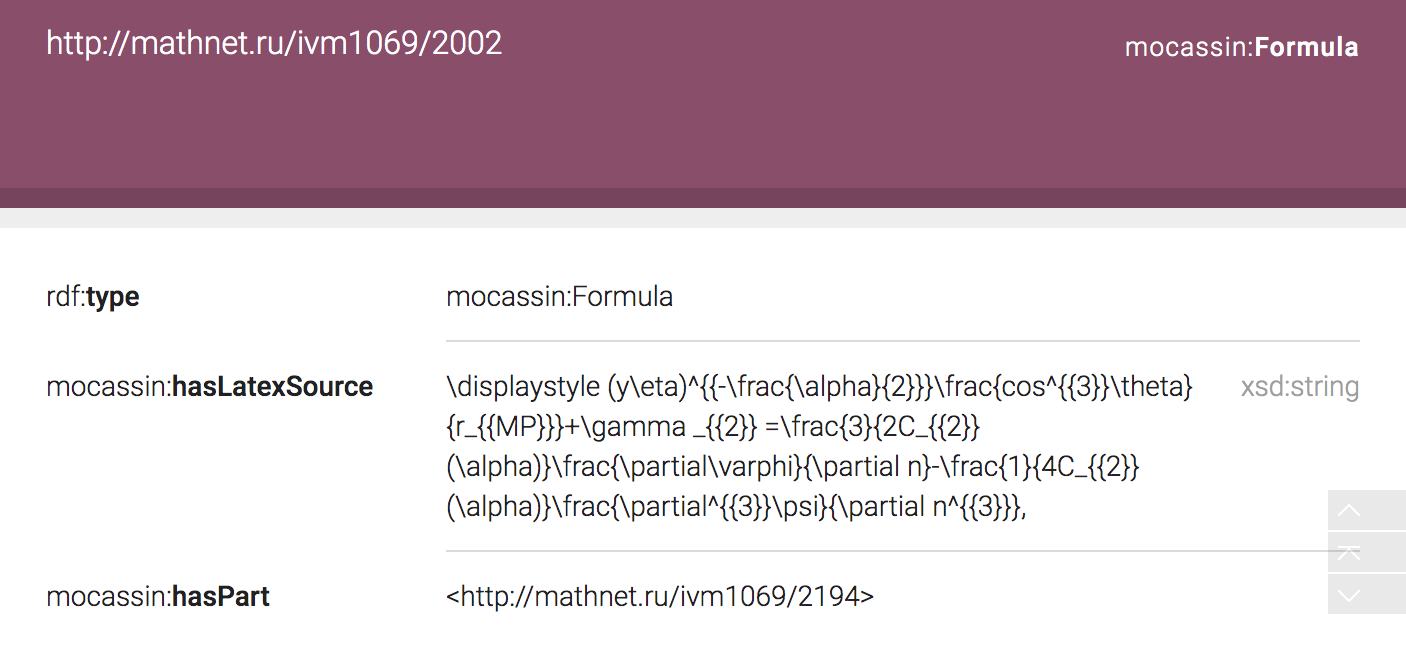}}
\caption{Visualization of math notation without rendering.} \label{fig:latex}
\end{figure}

\textbf{9) Support of \LaTeX\ math notation.} LOD datasets may contain literals with math notation in the \LaTeX\ format. LodView doesn’t render this notation and visualizes it just as its source \LaTeX\ code (see Fig. \ref{fig:latex}). LodView must be able to visualize math notation it the rendered form.

\section{Implementation of the improvements}

Currently, we have implemented the improvements \#\#1–4 and \#9, and partially the improvements \#5 and \#6.

\textbf{Issue \#1 (Resolution of Cyrillic URIs).} After analyzing the source code of LodView, we found that this problem is related to the fact that the URI text is represented in the \texttt{SPARQLEndPoint} class in Western ISO-8859-1 encoding. This problem was solved by adding a single line of code to the specified class that recodes the URIs from ISO-8859-1 to Unicode:

\texttt{IRI = new String (URI.getBytes ("ISO-8859-1"), "UTF-8");}

After we solved this problem, LodView began to correctly resolve Cyrillic URIs and generate a web page displaying the corresponding resource when accessing them. However, after this, we found a new problem: the resource URI is also displayed incorrectly on the returned web page. % (see Fig. \ref{fig:cyrillic-uris-new-problem}).

% \begin{figure}[ht]
% \frame{\includegraphics[width=\columnwidth]{fig-cyrillic-uris-new-problem.png}}
% \caption{Incorrect display of the resource URI.} \label{fig:cyrillic-uris-new-problem}
% \end{figure}

We solved this new problem by performing a similar recoding in the code of the corresponding web page (\texttt{resource.jsp}).

\textbf{Issue \#3 (Support of Cyrillic literals).} This problem was solved in the same way as the previous one, using a similar recoding.

\textbf{Issue \#2 (Decoding Cyrillic URIs in Turtle representations of resources).} To solve this problem, we have made changes to the \texttt{ResourceBuilder} class.

\textbf{Issue \#4 (User-friendly URLs for RDF representations of resources).} To solve this problem, we added support for such URLs by making changes to the \texttt{ResourceController} class.

\begin{figure}[t]
\frame{\includegraphics[width=\columnwidth]{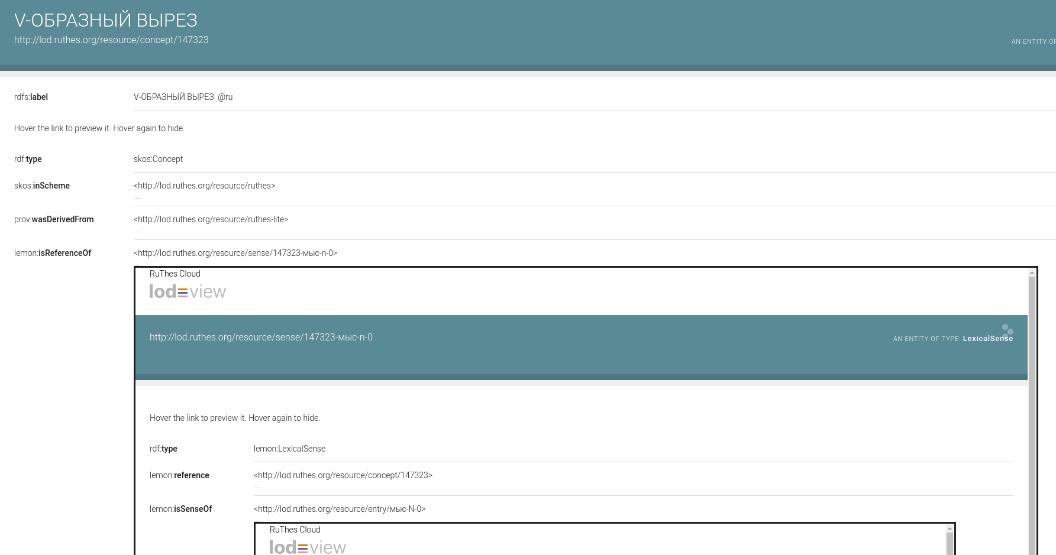}}
\caption{Example of deploying an embedded resource using an iframe.} \label{fig:nested-node-solved}
\end{figure}

\textbf{Issue \#6 (Expanding nested resources).} The problem was solved as follows. Near URIs of the nested resources, we added ``Expand'' buttons. When the user clicks on this button, an iframe with the representation of this nested resource is appended just near its URI (see Fig. \ref{fig:nested-node-solved}). Also, we are going to implement more flexible solution, that doesn’t use an iframe. 

\begin{figure}[t]
\frame{\includegraphics[width=\columnwidth]{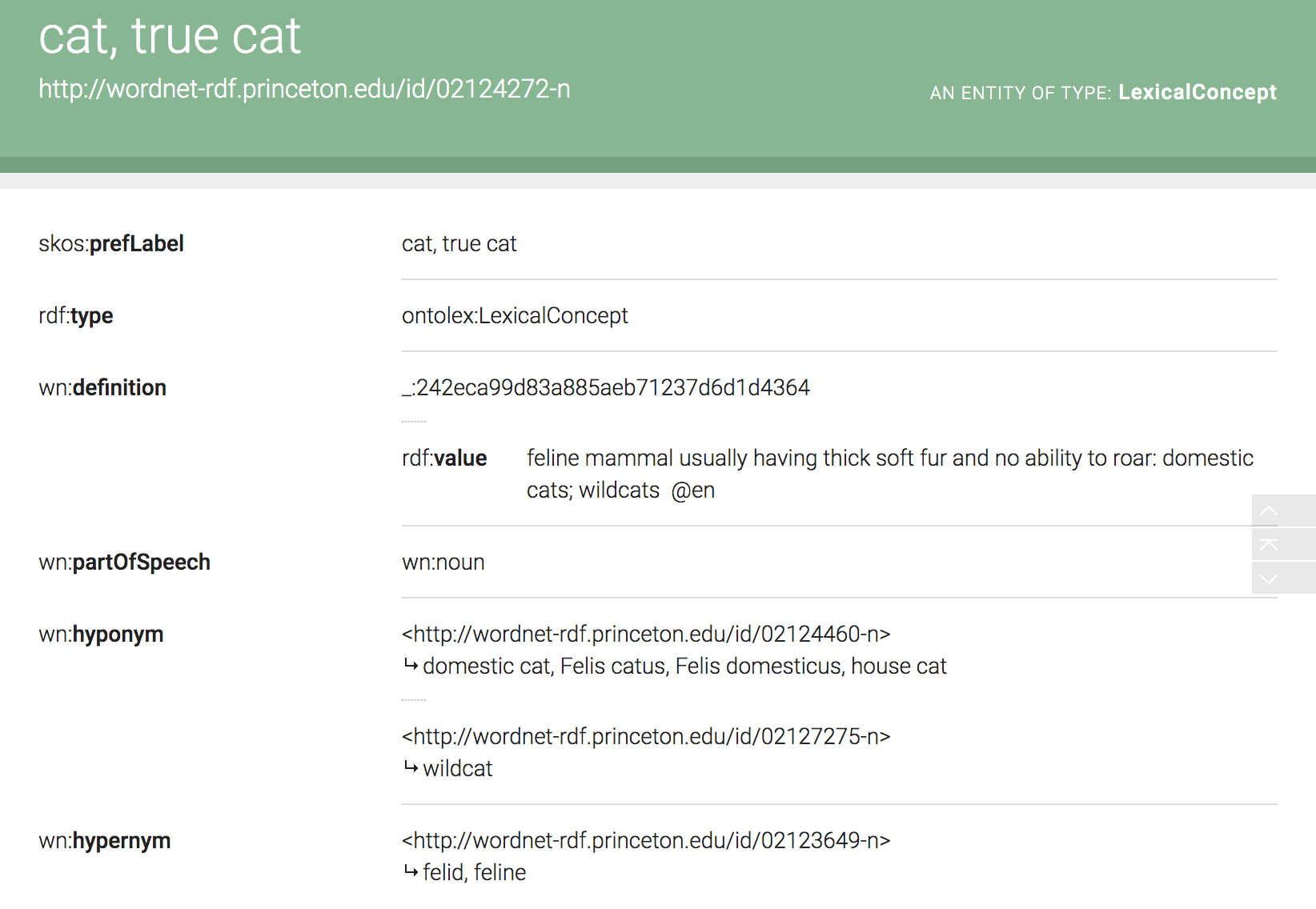}}
\caption{Blank node, located in the cell with the property value.} \label{fig:blank-node-solved}
\end{figure}

The descriptions of blank nodes are simply moved from the bottom of the HTML page to the cells with the IDs of the these nodes. For example, Fig. \ref{fig:blank-node-solved} represents the \textit{cat} synset. For this synset, the value of the \texttt{wn:definition} property is a blank node. This blank node is represented not only by its ID, but also by its description, i.e. a table of its own properties. 

\begin{figure}[h]
\frame{\includegraphics[width=\columnwidth]{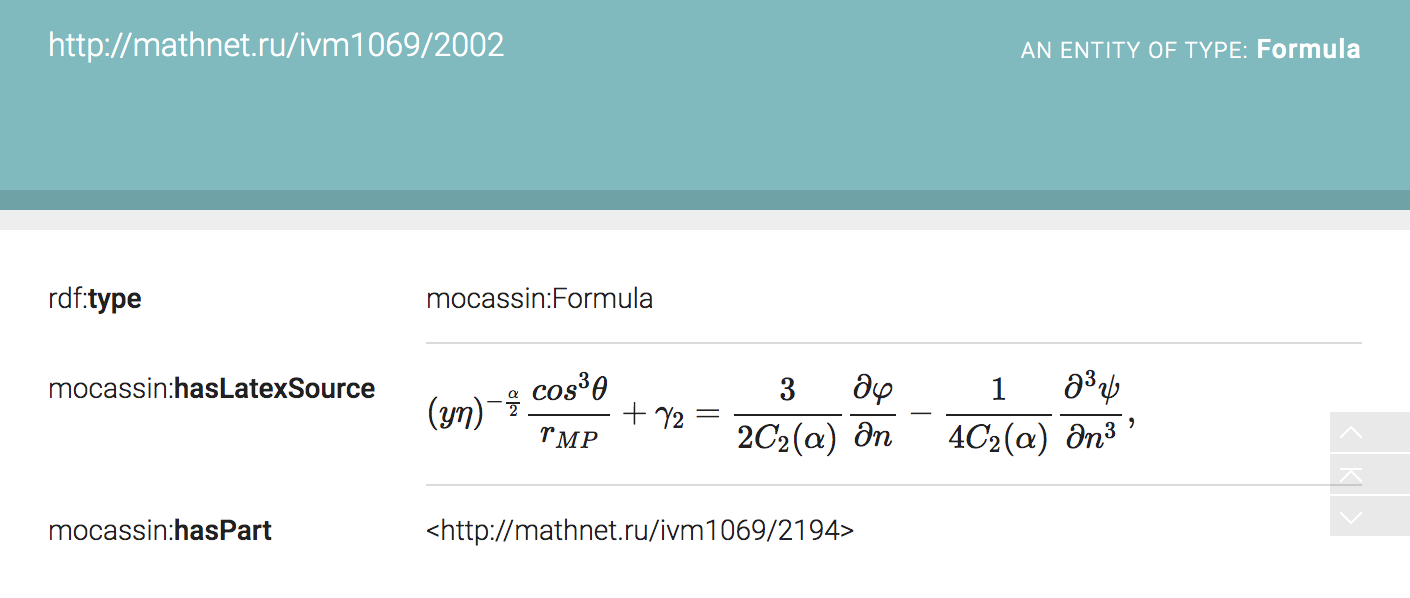}}
\caption{Visualization of math notation with rendering.} \label{fig:latex-solved}
\end{figure}

\textbf{Issue \#9 (Support of \LaTeX\ math notation).} In order to solve this problem, we have developed a special plugin, based on the well-known MathJax library\footnote{\url{https://www.mathjax.org/}} \cite{cervone2012} (see Fig. \ref{fig:latex-solved}).

\section{Conclusion}

We reveal several limitations of LodView that impede its use for navigation over the multilingual Linguistic Linked Open Data cloud. For fixing these limitations, we propose the following improvements: 1) resolution of Cyrillic URIs; 2) decoding Cyrillic URIs in Turtle representations of resources; 3) support of Cyrillic literals; 4) user-friendly URLs for RDF representations of resources; 5) support of hash URIs; 6) expanding nested resources; 7) support of RDF collections; 8) pagination of resource property values; and 9) support of \LaTeX\ math notation.

At the moment, we have implemented the improvements \#\#1–4 and \#9, and partially the improvements \#5 and \#6. Our fork of LodView is available on GitHub: https://github.com/ManlyMan1/LodView\_Cyrillic.

Currently, we are working on the remaining issues. After that, we are going to send a pull request to the main LodView repository.

\textbf {Acknowledgment.} The work was initiated for navigation over LLOD cloud at KFU where was funded by RSF according to the project no. 19-71-10056, and then adapted for navigation over specialized \LaTeX -based math collections at JSC RAS.

\bibliographystyle{unsrturl}
\bibliography{SETIT2022.bib}

% \begin{thebibliography}{00}
% \bibitem{b1} G. Eason, B. Noble, and I. N. Sneddon, ``On certain integrals of Lipschitz-Hankel type involving products of Bessel functions,'' Phil. Trans. Roy. Soc. London, vol. A247, pp. 529--551, April 1955.
% \bibitem{b2} J. Clerk Maxwell, A Treatise on Electricity and Magnetism, 3rd ed., vol. 2. Oxford: Clarendon, 1892, pp.68--73.
% \bibitem{b3} I. S. Jacobs and C. P. Bean, ``Fine particles, thin films and exchange anisotropy,'' in Magnetism, vol. III, G. T. Rado and H. Suhl, Eds. New York: Academic, 1963, pp. 271--350.
% \bibitem{b4} K. Elissa, ``Title of paper if known,'' unpublished.
% \bibitem{b5} R. Nicole, ``Title of paper with only first word capitalized,'' J. Name Stand. Abbrev., in press.
% \bibitem{b6} Y. Yorozu, M. Hirano, K. Oka, and Y. Tagawa, ``Electron spectroscopy studies on magneto-optical media and plastic substrate interface,'' IEEE Transl. J. Magn. Japan, vol. 2, pp. 740--741, August 1987 [Digests 9th Annual Conf. Magnetics Japan, p. 301, 1982].
% \bibitem{b7} M. Young, The Technical Writer's Handbook. Mill Valley, CA: University Science, 1989.
% \end{thebibliography}

\end{document}